\documentclass[letterpaper, 10 pt, conference]{ieeeconf}  

\IEEEoverridecommandlockouts                              

\usepackage[utf8]{inputenc}
\usepackage{makecell}
\usepackage{booktabs}
\usepackage{subcaption}
\usepackage{todonotes}
\usepackage[english]{babel}
\usepackage{dirtytalk}
\usepackage[hidelinks]{hyperref}
\usepackage{cleveref}
\usepackage{placeins}

\usepackage{url}

\usepackage[backend=bibtex,style=numeric, maxcitenames=1, maxbibnames=2]{biblatex}
\title{\LARGE \bf Towards Zero-Shot Terrain Traversability Estimation: Challenges and Opportunities}

\author{Ida Germann, Mark O. Mints and Peer Neubert
\thanks{All authors are with the Intelligent Autonomous Systems Group, Institute of Computational Visualisitics, University of Koblenz, Germany
        {\tt\small \{idagermann, mmints, neubert\}@uni-koblenz.de}}%
}

\begin{document}


\maketitle
\thispagestyle{empty}
\pagestyle{empty}

\begin{abstract}
Terrain traversability estimation is crucial for
autonomous robots, especially in unstructured environments
where visual cues and reasoning play a key role. While
vision-language models (VLMs) offer potential for zero-shot
estimation, the problem remains inherently ill-posed. To explore
this, we introduce a small dataset of human-annotated water
traversability ratings, revealing that while estimations are subjective, human raters still show some consensus. Additionally,
we propose a simple pipeline that integrates VLMs for zero-
shot traversability estimation. Our experiments reveal mixed
results, suggesting that current foundation models are not yet
suitable for practical deployment but provide valuable insights
for further research.

\end{abstract}

\section{Introduction}
\label{sec:introduction}

Consider a ground robot that needs to navigate through an unfamiliar forest.
Relying on geometric features alone may be insufficient when faced with obstacles such as tall grass or uneven surfaces \cite{christossevastopoulosSurveyTraversabilityEstimation2022}.
Using additional visual information could provide valuable semantic information about the terrain, but this still might be insufficient \cite{tianruiguanGANavEfficientTerrain2022}.
For instance, if the robot has learned that puddles are generally traversable, it might fail to recognize that a particular puddle is excessively muddy and thus unsafe.
Conversely, if the robot has been
trained to avoid water, it might miss an opportunity to cross
a shallow puddle that is safer than navigating through dense
vegetation. Adopting a dynamic approach to handle these
variations would significantly enhance the robot’s decision-
making capabilities in real-world scenarios.

Intuitively, we humans are able to estimate traversability
of different terrains based on visual cues and common-sense
reasoning -- e.g. by recognizing that a shallow looking puddle
on sturdy ground is likely traversable, while a deep puddle
on soft ground is not. Recent advancements in foundation
models suggest that robots might achieve similar reasoning
capabilities. However, the problem remains ill-posed, as
traversability depends not only on visible cues but also on
hidden factors like water depth and surface consistency.
Additionally, human annotations of traversability vary due to
subjective interpretations, similar to challenges in semantic
segmentation \cite{antonelliViewComputationalModels2022}.

To investigate this, we introduce a dataset of water
traversability ratings and propose a straightforward pipeline
concept using vision-language models (VLMs) for zero-shot
terrain traversability estimation.

\section{Zero-Shot Terrain Traversability Estimation -- An ill-posed Problem}
\label{sec:problem}

In the following, a small dataset on the traversability of
water instances is introduced, initially created to evaluate
foundation models in our pipeline (Sec. \ref{sec:concept}). Beyond this, the dataset provides key insights into the inherent ill-posedness of zero-shot terrain traversability estimation.

The dataset comprises 195 images, including subsets from RUGD \cite{maggiewignessRUGDDatasetAutonomous2019} and RELLIS-3D \cite{pengjiangRELLIS3DDatasetData2021}, as well as self-taken and CC-0 online images. It contains 254 water instances, each labeled
for the two robot types Clearpath Husky A200 \cite{ClearpathHuskyA200} and Unitree B1 \cite{UnitreeB1}, resulting in 508 annotated instances in total.
Seven annotators assigned ratings based on their estimation
of whether a robot could traverse a given water body without
getting stuck or damaged. The rating scheme thereby is based
on \cite{tianruiguanGANavEfficientTerrain2022}: 1 – smooth, 2 – rough, 3 – bumpy, 4 – non-navigable/forbidden.

\begin{figure}[htbp]
    \centering
    \includegraphics[width=0.85\linewidth]{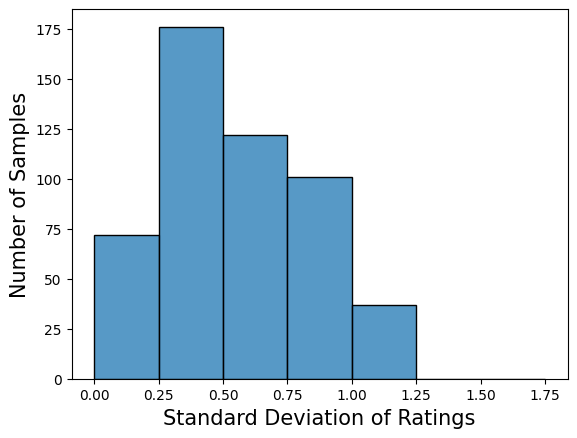}
    \caption{Distribution of the standard deviations of the annotators' ratings.
    Num. rated instances per annotator: 508. 
    }
    \label{fig:annotation_stds}
\end{figure}

Figure \ref{fig:annotation_stds} shows the distribution of standard deviations in
these annotations. While varying degrees of disagreement
are observed across instances, the majority exhibit standard
deviations below 1.0, indicating a reasonable level of consensus among annotators, despite the problem’s ambiguity.
However, subjective interpretations and the absence of standardized criteria introduce noise into the dataset.
\begin{figure*}[htbp]
    \centering
    \includegraphics[width=0.95\linewidth]{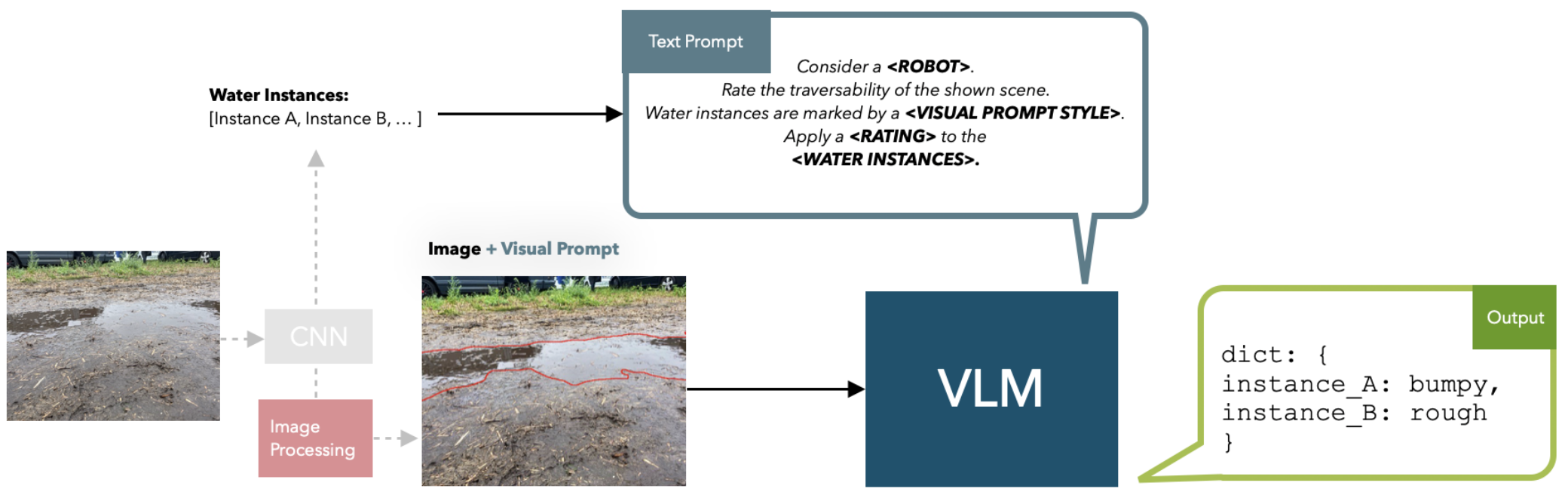}
    \caption{Illustration of the proposed pipeline concept. The CNN-block in light gray represents an optional modification to extract water instances. In our experiments, manually annotated ground truth masks are used instead.
    }
    \label{fig:pipeline_overview}
\end{figure*}

The lack of a unified rating framework further complicates
the problem. Existing studies \cite{tianruiguanGANavEfficientTerrain2022,davidd.fanSTEPStochasticTraversability2021, elnoorAMCOAdaptiveMultimodal2024} use differing definitions of traversability, making direct comparisons difficult.
Future experiments should explore strategies to reduce this
inconsistency, e.g. by refining annotation guidelines and
incorporating expert raters that have in-field experience with
the considered robot types. Additionally, designing a larger
and more diverse dataset with varied environmental conditions and robot-specific constraints could allow for more
robust evaluations.

\section{Pipeline Concept}
\label{sec:concept}

Our proposed pipeline provides a simple yet flexible
framework for integrating vision-language models (VLMs)
for zero-shot terrain traversability estimation.
Unlike previous methods that rely on predefined traversability mappings
\cite{tianruiguanGANavEfficientTerrain2022}, our approach uses a vision-language model (VLM) to allow for context-aware reasoning.

Figure \ref{fig:pipeline_overview} illustrates the pipeline’s core components. It
takes an RGB image as input and yields a dictionary of
traversability ratings for each water instance as output. The
dictionary then can be used to update a segmentation mask,
producing a context-aware traversability cost-map.

In detail, the pipeline consists of the following steps: First,
water instances are extracted from the input image using a
semantic segmentation mask. For prototype testing, we use
ground-truth segmentation masks, but real-world deployment
could rely on models such as FastSAM \cite{zhaoFastSegmentAnything2023}.
The extracted
instances, along with a structured text prompt describing the
robot type, task, and rating scheme, are fed into a VLM,
which assigns traversability ratings. These ratings can be
used to update the segmentation mask, producing a context-
aware traversability cost-map for the robot.

For on-device deployment, we integrate small and quantized versions of state-of-the-art VLMs: LLaVA1.6-7B\footnote{\url{https://ollama.com/library/llava:7b}} (7 billion parameters), LLaVA-Phi3-Mini
\footnote{\url{https://ollama.com/library/llava-phi3}} (4 billion parameters), and MiniCPM-V 2.6
\footnote{\url{https://ollama.com/library/minicpm-v}} (8 billion parameters), all quantized to 4-bit integer precision.
Additionally, we evaluate GPT-4o via the OpenAI API\footnote{\url{https://platform.openai.com/docs/models} (model-tag 'gpt-4o-2024-08-06'), all accessed 4th Feb., 2025} to compare performance against a larger-scale model.

\section{Experimental Results}
\label{sec:insights}
For evaluating the performance of different VLMs in the introduced pipeline framework, we conducted experiments on the water traversability dataset, considering the traversability estimation task as a multi-class classification problem.
While the small VLMs showed a basic ability to reason about traversability, their predictions were particularly sensitive to prompt variations and temperature settings, leading to inconsistent results.
Implemented prompting techniques were, e.g., \say{rephrase-and-respond} \cite{dengRephraseRespondLet2024}, \say{role prompting} \cite{wangRoleLLMBenchmarkingEliciting2024}, and \say{chain-of-thought (CoT) reasoning} \cite{weiChainofThoughtPromptingElicits2023}.

GPT-4o showed improved generalization but still suffered from absolute performance limitations on the evaluated dataset.
While the poor F1-scores on any class of at most 0.51 could be partly caused by the ill-posedness of the visual traversability estimation problem itself, the overall performance of the models suggests that they are not yet suitable for practical applications in this domain.
Depending on the used textual prompting strategy, the models partly suffered from a poor task understanding, which could even lead to entirely failed predictions -- e.g. no dictionary output, but some unsuitable text response.

\section{Conclusion}
\label{sec:conclusion}

Our study highlights the challenges of zero-shot terrain
traversability estimation, particularly the ambiguity in human
annotations and the absence of standardized rating criteria.
While our experiments indicate that current vision-language
models struggle with consistent and reliable predictions,
the observed partial agreement among human annotators
suggests that the ill-posed problem of zero-shot traversability
estimation is approachable and therefore worth further exploration. The proposed pipeline provides a simple yet flexible
framework for further research of foundation models in this
domain. Future work should moreover focus on improving
dataset quality and establishing standardized rating criteria
to enhance consistency in terrain traversability estimation.


\setcounter{biburlucpenalty}{1000}
\printbibliography[title={References}]

\end{document}